\pdfoutput=1

\documentclass[11pt]{article}

\usepackage{acl}

\usepackage{times}
\usepackage{latexsym}

\usepackage{CJKutf8}

\usepackage[T1]{fontenc}

\usepackage[utf8]{inputenc}

\usepackage{microtype}

\usepackage{graphicx}
\usepackage{lingmacros}

\usepackage{multirow}
\title{Generate, Evaluate, and Select: A Dialogue System with a Response Evaluator for Diversity-Aware Response Generation}

\author{Ryoma Sakaeda \and Daisuke Kawahara \\
        Waseda University \\
        \texttt{\{s.ryoma6317@akane., dkw@\}waseda.jp}}

\begin{document}
\maketitle
\begin{abstract}
We aim to overcome the lack of diversity in responses of current dialogue systems and to develop a dialogue system that is engaging as a conversational partner. We propose a generator-evaluator model that evaluates multiple responses generated by a response generator and selects the best response by an evaluator. By generating multiple responses, we obtain diverse responses. We conduct human evaluations to compare the output of the proposed system with that of a baseline system. The results of the human evaluations showed that the proposed system's responses were often judged to be better than the baseline system's, and indicated the effectiveness of the proposed method.
\end{abstract}

\section{Introduction}

Dialogue systems based on deep neural networks (DNNs) have been widely studied. Although these dialogue systems can generate fluent responses, they often generate dull responses such as ``yes, that's right'' and lack engagingness as a conversation partner~\cite{jiang-de-rijke-2018-sequence}. To develop an engaging dialogue system, it is necessary to generate a variety of responses not to bore users.

However, dialogue systems that are capable of generating diverse responses are difficult to automatically evaluate. A commonly used evaluation metric is BLEU~\cite{papineni-etal-2002-bleu} used in machine translation, which measures the degree of n-gram agreement with the reference response. However, due to the diversity of responses, i.e., the one-to-many nature of dialogue~\cite{zhao-etal-2017-learning}, which means the existence of multiple appropriate responses to an utterance, methods that compare the response to reference responses are not appropriate. Therefore, there is a need for evaluation methods that do not use reference responses, and one of them is supervised evaluation. It trains DNNs using human evaluations of responses generated by humans and models~\cite{zhao-etal-2020-designing,ghazarian-etal-2019-better}. DNN-based evaluations correlate to some extent with human evaluations. 

We aim to develop a dialogue system that is more engaging as a conversational partner by combining  independently studied response generation and response evaluation models into a single dialogue system. Specifically, we propose a generator-evaluator model in which multiple responses are generated by the generation model, evaluated by the evaluation model, and  the response with the highest evaluation score is selected. By generating multiple responses, we can obtain diverse responses. This can be enabled by the response evaluator that does not require reference responses.

Our methods of generating multiple responses include a method with multiple decoding schemes and a method that uses a model that can generate responses with a specified Dialogue Act (DA). Generating responses by specifying various DAs leads to a variety of responses.

To evaluate the proposed method, we conducted human evaluation by crowdsourcing to compare the outputs of the proposed system and a baseline system. The evaluation results show that the proposed system outputs better responses, and  indicate the effectiveness of the proposed method.

We target Japanese dialogue systems and construct datasets of Japanese dialogues.

\begin{figure*}
  \centering
  \includegraphics[width=.8\linewidth]{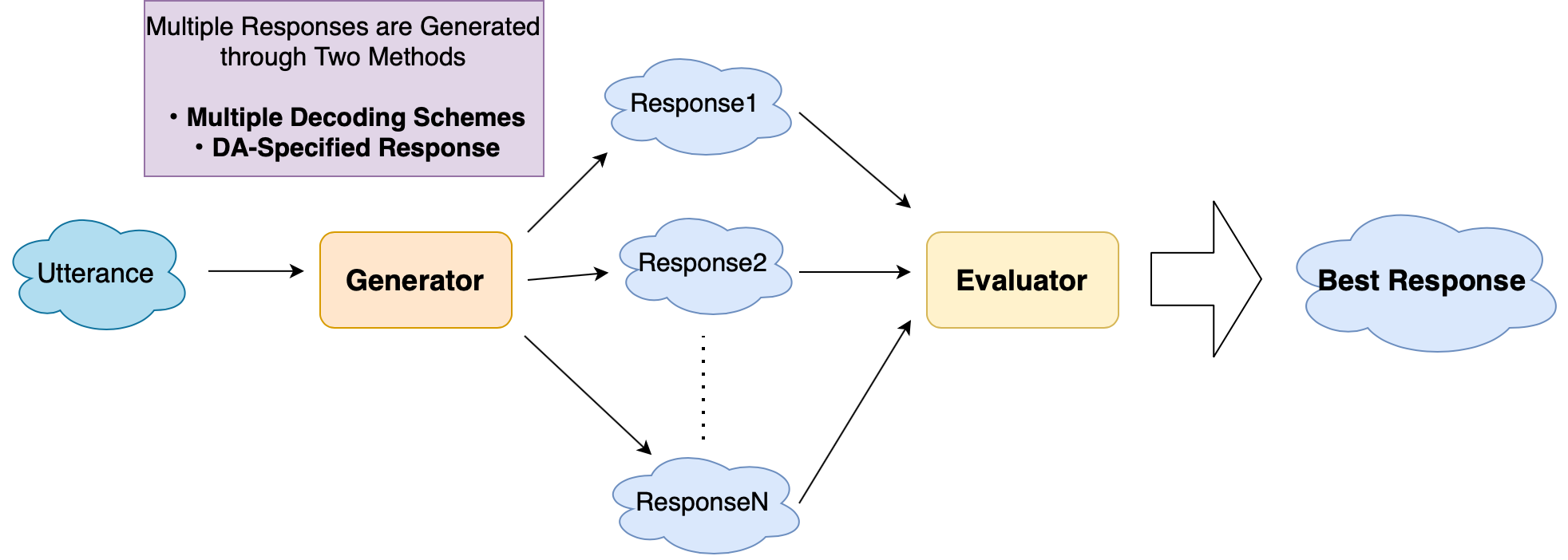}
  \caption{The architecture of our proposed system, the generator-evaluator model. It generates multiple responses from the generator, evaluates them with the evaluator, and selects the best response.}
  \label{graph:system}
  \vspace{-10pt}
\end{figure*}

\section{Related Work}
\label{sec:related}
Methods for evaluating responses by dialogue systems can be divided into human and automatic evaluations. Automatic evaluation can be further classified into evaluation with or without reference responses.
As an automatic evaluation metric, BLEU~\citep{papineni-etal-2002-bleu} is mainly used. It evaluates responses in terms of n-gram agreement with the reference sentence. However, it has been shown that there is no correlation at all between BLEU and human evaluations \citep{liu-etal-2016-evaluate}. One reason for this is the one-to-many nature of dialogue~\citep{zhao-etal-2017-learning}, which means that there are multiple appropriate responses to an utterance. Considering this nature, a method that measures the degree of n-gram agreement with the reference response is inappropriate for evaluating responses. Therefore, automatic evaluation methods without any reference responses have been studied~\citep{zhao-etal-2020-designing,ghazarian-etal-2019-better}. They trained BERT~\citep{devlin-etal-2019-bert} on a dataset of human evaluations to perform response evaluation that correlates with the human evaluations.
\\ \indent DA represents the role of an utterance in a dialogue. There are some datasets annotated with DAs such as SwDA~\citep{stolcke-etal-2000-dialogue} and MRDA~\citep{shriberg-etal-2004-icsi}. However, such datasets exist only for English, and we construct a DA dataset in Japanese. ~\citet{raheja-tetreault-2019-dialogue,10.1145/3331184.3331375} constructed a model that classifies a DA for an utterance. \citet{kawano-etal-2019-neural}~proposed a model to generate responses with a specified DA. This was achieved through adversarial learning. In this study, we use a more straightforward method to control responses.


\begin{table*}[t]
    \centering
    \small
    \begin{tabular}{l|l|r} \hline
        Viewpoint & Response & Amount \\\hline \hline
        Relevance & Twitter/decoding model & 4,000/4,000 \\
        Interestingness & Twitter & 2,000 \\
        Engagingness & Twitter/decoding model/DA model & 4,000/4,000/4,000 \\
        Empathy & Twitter & 2,000 \\\hline
    \end{tabular}
    \caption{Amount of data for each viewpoint in the Response-Evaluation dataset. "Response" indicates where the response derives from. Due to the collection cost, more data were collected for the more important viewpoints.}
    \label{tbl:evaluation-dataset}
\end{table*}

\begin{table*}[t]
\centering
\small
\begin{tabular}{l|l} \hline
     Dialogue Act&Description   \\\hline \hline
     Advice&  advice or instruction given to the partner\\ 
     Emotion & emotion experienced by speaker\\
     Opinion & opinion about a particular topic\\
     Inform & give information about oneself(speaker) \\
     Schedule & what the speaker plans to do or wants to do\\
     Question & questioning the partner\\
     Agree & agree about the partner's opinion or feeling\\\hline
\end{tabular}
\caption{DA types and their descriptions. Crowdworkers are shown this description and asked to choose which DA applies to each response.}
\label{tbl:da}
\end{table*}

\section{A Generator-Evaluator Model for an Engaging Dialogue System}
\label{sec:method}

\subsection{Generator-Evaluator Model}
\label{ssec:method:generator-evaluator}

We propose a generator-evaluator model that generates multiple responses, evaluates these responses, and selects the response with the highest evaluation score for output. The overview of the proposed model is shown in Figure~\ref{graph:system}. Two methods are used to generate multiple responses: multiple decoding schemes and a model that can generate DA specified responses.
For the evaluator, BERT is fine-tuned with the Response-Evaluation dataset described in Section \ref{ssec:dataset:evaluation}.

\subsection{Multiple Response Generators}
\label{ssec:method:multi-response}
We use T5~\cite{2020t5}    as a generator by fine-tuning it with the method described below.
\subsubsection{Multiple Decoding Schemes}
\label{sssec:method:multi-response:decoding-scheme}
The first method for obtaining multiple responses is to use multiple decoding schemes. Three types of decoding methods are used: greedy search, beam search, and sampling. In particular, to repeat sampling is thought to generate diverse responses. We use the top-50 sampling~\citep{fan-etal-2018-hierarchical}.

\subsubsection{DA-Specified Response Generation}
\label{sssec:method:multi-response:da}
The second method to obtain multiple responses is to use a model that can generate responses with specified DAs. We achieve such a model by training a response generation model based on utterance-response pairs attached with prompts that specify the DA of a response. The dataset format is as follows: (\ref{example:utterance}) represents the input and (\ref{example:response}) represents the response. The italic span denotes the prompt specifying a DA.

\eenumsentence{
    \item \label{example:utterance}
   \textit{Return a response of advice to the interlocutor} I haven't done the assignment yet. 
    \item \label{example:response}
You should read this book before you do it.
}

To train this model, we need a dialogue corpus annotated with DA labels. We use the DA dataset described in Section \ref{ssec:dataset:da}. A dialogue corpus without DA labels is also used as responses with a \textit{general} DA. Its prompt is \textit{Return a response}.
\begin{table}[t]
\centering
\small
\begin{tabular}{l|r} \hline
     Dialogue Act& Amount   \\\hline \hline
     Advice& 853 \\ 
     Emotion & 1,433 \\
     Opinion &1,323 \\
     Inform &  1,131\\
     Schedule & 718\\
     Question & 342\\
     Agree & 1,136\\\hline
\end{tabular}
\caption{Amount of data for each DA.}
\label{tbl:da-dataset}
\end{table}

\section{Dataset}
\label{sec:dataset}
Since there is not a sufficiently large corpus of Japanese dialogues, we start from corpus construction.

\subsection{Twitter Dataset}
\label{ssec:dataset:twitter}
Our dialogue dataset is collected from Twitter using the Twitter API. Some of the conversations are collected from single-turn conversations only~(Twitter-Single), while the others are collected from multi-turn conversations~(Twitter-Multi).

\subsection{Response-Evaluation Dataset}
\label{ssec:dataset:evaluation}

Our Response-Evaluation dataset contains evaluations of how well a response meets certain viewpoints when looking at a single-turn utterance and response. We use the following four evaluation viewpoints: relevance, interestingness, engagingness, and empathy.

We use two types of utterance-response pairs to ensure corpus diversity: the first is the Twitter-Single dataset described in Section \ref{ssec:dataset:twitter}, and the second is the utterances from the Twitter-Single dataset and the responses generated from generator models. We use two types of generator models: the model with the multiple decoding schemes and the model that can generate responses with specified DAs. In the datasets using responses from the generator models, the evaluations of multiple responses to an utterance are collected. They represent how evaluations differ when different responses are generated to the same utterance. The evaluations are collected through crowdsourcing. We ask a five-grade question to five people, and the average was taken as the evaluation value. The statistics of the dataset is shown in Table~\ref{tbl:evaluation-dataset}.

\subsection{DA Dataset}
\label{ssec:dataset:da}

We assign DAs for each utterance in the Twitter-Multi dataset described in Section \ref{ssec:dataset:twitter}. By using the dataset of multi-turn conversations, we intended to make a dataset to capture the transition of DAs in a long conversation. We adopt seven DA types shown in Table \ref{tbl:da}. The number of DA types was reduced to seven because the 42 types in the previous study \cite{stolcke-etal-2000-dialogue} were too fine-grained to be annotated by crowdsourcing. Since there are utterances that do not settle on a single DA, we allow multiple DAs for each utterance. DAs are collected through crowdsourcing. We ask a question to five people and adopt the DA with at least three votes. The amount of utterances for each DA is shown in Table \ref{tbl:da-dataset}. Since the amount of data is not sufficient to be used for training the generator model described in Section \ref{sssec:method:multi-response:da}, this dataset is used to train DA classifiers that are applied to the Twitter-Single dataset for data augmentation.

\begin{table}[t]
\centering
\small
\begin{tabular}{l|r|r|r} \hline
     Dialogue Act& Precision & Recall & F1   \\\hline \hline
     Advice& 0.52 & 0.57 & 0.54 \\ 
     Emotion & 0.54 & 0.37 & 0.44 \\
     Opinion & 0.60 & 0.51 & 0.55\\
     Inform & 0.44 & 0.55 & 0.49 \\
     Schedule & 0.41 & 0.47  & 0.44\\
     Question & 0.88 & 0.51 & 0.65\\
     Agree & 0.69 & 0.53 & 0.60 \\\hline
\end{tabular}
\caption{Results of DA classification by five-fold cross validation.}
\label{tbl:da-classifier}
\end{table}

\begin{table}[t]
\centering
\small
\begin{tabular}{l|r} \hline
     Dialogue Act& Amount   \\\hline \hline
     Advice& 2,284 \\ 
     Emotion & 4,195 \\
     Opinion & 6,580\\
     Inform &  63,652\\
     Schedule & 89,990\\
     Question & 33,629\\
     Agree & 70,557\\\hline
\end{tabular}
\caption{Amount of data for each DA obtained by data augmentation with the DA classifiers.}
\label{tbl:da-augmentation}
\end{table}

\subsubsection*{Augmentation with DA Classifiers}
\label{sssec:dataset:da:classifier}
   
We build DA classifiers by fine-tuning BERT with the DA dataset described above. These DA classifiers are binary classifiers that determine whether a response belongs to each of the  DAs. The results of DA classification by each DA classifier are shown in Table \ref{tbl:da-classifier}. Metrics are precision, recall, and F1. They are computed using five-fold cross validation. From this table, the predicted DAs do not seem sufficiently precise to
be used for data augmentation. However, we manually examined a part
of predicted DAs and found that their precision was around 70\%, which
made us decide to use them for data augmentation.

We augment the DA dataset by applying the classifiers to an unlabeled dialogue corpus. We apply each binary classifier to 1.6M responses of the Twitter-Single dataset, and assign DA labels to responses judged to be positive. The amount of data obtained for each DA is shown in Table \ref{tbl:da-augmentation}.


\begin{table}[t]
    \small
       \begin{tabular}{l||r|r|r} 
    \hline
       Comparison & Win & Lose & Even \\\hline \hline 
       DE Best vs DE Greedy & \textbf{44\%} & 21\% & 35\% \\
        DE Best vs DE Random & \textbf{50\%} & 24\% & 26\% \\
        DA Best vs DA General & \textbf{42\%} & 25\% & 33\% \\
        DA Best vs DA Random & \textbf{44\%} & 21\% & 35\% \\\hline 
        DADE Best vs DE Greedy & \textbf{44\%} & 43\% & 12\% \\
        DADE Best vs DE Random & \textbf{48\%} & 41\% & 11\% \\
        DADE Best vs DA General & \textbf{49\%} & 33\% & 17\% \\
        DADE Best vs DA random & \textbf{55\%} & 28\% & 17\% \\
        DADE Best vs DADE Random & \textbf{73\%} & 14\% & 13\% \\\hline
        DADE Best vs DE Best & 38\% & \textbf{51}\% & 11\% \\
        DADE Best vs DA Best & \textbf{45}\% & 32\% & 22\% \\\hline
    \end{tabular}
    \caption{Result of one-to-one comparison between a proposed system and a baseline system.}
    \label{tbl:result}
\end{table}

\begin{table*}[t]
    \small
    \centering
    \begin{tabular}{p{8mm}p{50mm}p{12mm}p{60mm}p{12mm}} \hline
        System & Utterance & Scheme/DA & Response &Eval \\\hline \hline
        DE &\begin{CJK}{UTF8}{ipxm}美味しいですよね。台湾茶。大好き。\end{CJK} (It's delicious. Taiwanese tea. I love it.) & Beam & \begin{CJK}{UTF8}{ipxm} ありがとうございます。台湾茶は美味しいですよね。 \end{CJK}(Thank you. Taiwanese tea is delicious.) & 3.6\\\cline{3-5}
        & & Sampling1 & \begin{CJK}{UTF8}{ipxm} 美味しいですよね。味も飲みやすくて、お値段の割にちょっぴり高級感があって良いですね(\verb|^|\verb|-|\verb|^|)
\end{CJK}(It's delicious. It tastes easy to drink and has a little bit of luxury for the price~(\verb|^|\verb|-|\verb|^|)~.)& 3.9 \\\cline{3-5}
        &&Sampling2&\begin{CJK}{UTF8}{ipxm} 台湾茶は美味しいですよねー。私も大好きです〜。
 \end{CJK}(Taiwanese tea is delicious! I love it too!) & 3.8 \\\hline
        DA & \begin{CJK}{UTF8}{ipxm} 充電がすぐ切れるんだね。 \end{CJK}(I see that the charge runs out quickly.) & General & \begin{CJK}{UTF8}{ipxm} 充電がすぐ切れるのね。 \end{CJK}(You see the charge runs out quickly, don't you.) & 3.4 \\ \cline{3-5}
         & & Advice &\begin{CJK}{UTF8}{ipxm} 充電が切れるなら、バッテリー交換した方がいいよ。 \end{CJK} (If it's going to run out of charge, you should replace the battery.) & 3.3 \\ \cline{3-5}
         & & Opinion & \begin{CJK}{UTF8}{ipxm} 
         充電切れたので、今日はお休みにします。
         \end{CJK}(I'm out of charge, so I'm taking the rest of the day off.) & 3.1  
         \\\hline
    \end{tabular}
    \caption{Diverse responses from the proposed system. We think that the best responses for humans are the responses of ``Sampling1'' for DE and ``Advice'' for DA. This best response gets the highest evaluation in DE. However, this is not the case in DA. This may be one reason why the experimental results for DA are inferior to one for DE.}
    \label{tbl:system-response}
\end{table*}

\section{Experiments}
\label{sec:experiment}
We do the evaluation by crowdsourcing. Workers are shown the outputs of the two systems and asked which of the system they would prefer to continue the conversation with. We ask a question to three workers and take a majority vote as the result. The test corpus consists of 2,000 sentences from the Twitter-Single dataset described in Section \ref{ssec:dataset:twitter} which are not used for training.

\subsection{Experimental Setup}
\label{ssec:experimental-setup}

The proposed systems use two types of generators: one by the multiple decoding schemes~(\textbf{DE}) and one by DA specified responses~(\textbf{DA}). Also, by combining DE and DA, the DA generator can generate responses using the multiple decoding schemes~(\textbf{DADE}). We define \textbf{DE Best}, \textbf{DA Best}, and \textbf{DADE Best}, which refer to the response judged to be the best among multiple responses by the evaluators in DE, DA, and DADE, respectively. Here, in DE, seven responses were generated by repeating sampling five times in addition to greedy search and beam search. In DA, seven responses were obtained by generating responses for the general DA and excluding the emotion DA, whose classifier did not perform accurately. Multiple DAs were allowed for dataset construction, but only one DA was specified for generation. In DADE, seven responses are obtained for each of the seven DAs, resulting in a total of 49 responses. We perform a one-to-one comparison of each proposed system's response with the baseline system's response following~\citet{roller-etal-2021-recipes}. There are five types of responses to be compared, which are shown below.
\begin{description}
    \item[DE Greedy]
    a response generated by greedy search
    \item[DE Random]
    a randomly selected response from seven responses
    \item [DA General]
    a response generated by specifying the general DA
    \item[DA Random]
    a randomly selected response from seven DAs responses
    \item[DADE Random]
    a randomly selected response from 49 responses
\end{description}





\subsection{Training}
\label{ssec:training}
We use T5~\cite{2020t5} pretrained with a Japanese corpus\footnotemark as a generator in DE. We fine-tune it with 800,000 pairs from the Twitter-Single dataset described in Section \ref{ssec:dataset:twitter}. The generator model used in DA is further fine-tuned from the DE generator model with the augmented DA dataset in Section  \ref{ssec:dataset:da} and a part of the Twitter-Single dataset as general DA responses. It has the same size as the augmented DA dataset (270,000 pairs).

The evaluator is a fine-tuned BERT model and constructed for each of DE and DA. The dataset used for fine-tuning is the Engagingness data of the Response-Evaluation dataset described in Section \ref{ssec:dataset:evaluation}. It consists of 4,000 pairs derived from Twitter and 4,000 pairs from either of the DE and DA generators. For DADE, we use the same evaluator as DA.

\subsection{Result}
\label{ssec:result}
The evaluation results of our experiments are shown in Table \ref{tbl:result}. It shows the effectiveness of generating multiple responses and selecting the best response by the evaluator. However, the results of \textbf{DADE Best vs DE Greedy} and \textbf{DADE Best vs DE Best} show the responses of the DA generator were not rated better than the responses of the DE generator. This can be attributed to the fact that the distribution of the dataset was skewed by data augmentation, and further study is needed. Example responses generated by the proposed system are shown in Table~\ref{tbl:system-response}.

\footnotetext{{https://huggingface.co/sonoisa/t5-base-japanese}}

\begin{table}[t]
    \centering
    \small
    \begin{tabular}{l||r|r|r} 
    \hline
       Comparison & Win & Lose & Even \\\hline \hline 
       DE Best' vs DE Greedy & 47\% & 24\% & 28\% \\
        DE Best' vs DE Random & 47\% & 27\% & 26\% \\
        DA Best' vs DA General & 36\% & 25\% & 40\% \\
        DA Best' vs DA Random & 45\% & 25\% & 30\% \\\hline
    \end{tabular}
    \caption{One-to-one comparison between a proposed system with an OOD evaluator and a baseline system.}
    \label{tbl:ood}
    \vspace{-8pt}
\end{table}

\begin{table}[t]
    \centering
    \small
    \begin{tabular}{l|r}
    \hline
    Decoding Scheme     & Ratio \\\hline \hline
    Greedy-Search     & 12\% \\
    Beam-Search &  15\% \\
    Sampling (x5) & 73\% \\
    \hline
    \end{tabular}
     \caption{Analysis of which decoding scheme is selected. Sampling was repeated five times, and the percentage of any of the five responses chosen was 73\%.}
    \label{tbl:which-response-de}
    \vspace{-12pt}
\end{table}

\begin{table}[t]
    \centering
    \small
    \begin{tabular}{l|r}
    \hline
    DA     & Ratio \\\hline \hline
    General     & 16\% \\
    Advice &  8\% \\
    Schedule & 16\% \\
    Question & 11\% \\
    Inform & 14\% \\
    Agree & 9\% \\
    Opinion & 25\% \\
    \hline
    \end{tabular}
    \caption{Analysis of DA selection.}
    \label{tbl:which-response-da}
    \vspace{-12pt}
\end{table}

\section{Analysis}
\label{sec:ablation}


\subsection{Out-of-Domain Evaluator}
 In the experiments in Section \ref{sec:experiment}, each evaluator of DE and DA was trained using the human evaluations of the corresponding generator responses for each of DE and DA. However, it is not practical to use human evaluations for each generator. Therefore, we investigate the impact of using different generation methods and datasets used for evaluators. The same comparisons are made as the comparisons in Section \ref{sec:experiment}. The results are shown in Table \ref{tbl:ood}. We see that the proposed systems defeat the baseline in this case as well.


\subsection{Which Response is Chosen?}
\label{ssec:ablatioon:which-response}
We analyzed which decoding methods or DAs are selected by the evaluator model. The more equally the choices are divided, the more effective the proposed method is. This is because the proposed method cannot be surpassed by using any one specific decoding scheme or DA. The results of the analysis are shown in Tables \ref{tbl:which-response-de} and \ref{tbl:which-response-da}. The choices are scattered, and thus the proposed method can generate diverse responses.
\section{Conclusion}
\label{sec:conclusion}
We developed a dialogue system that can generate engaging responses by incorporating a response evaluator within the dialogue system. We proposed a generator-evaluator model, which consists of multiple response generation through multiple decoding schemes or specified DAs, responses evaluations, and the best response selection. Human evaluation showed that responses generated by the generator-evaluator model are more engaging than those by the baseline systems. However, it is still necessary to improve the quality of responses generated with specified DAs in the future.


\section*{Acknowledgements}
This work was supported by a joint research grant from LINE Corporation.

\bibliography{anthology,custom}

\begin{thebibliography}{15}
\expandafter\ifx\csname natexlab\endcsname\relax\def\natexlab#1{#1}\fi

\bibitem[{Ahmadvand et~al.(2019)Ahmadvand, Choi, and
  Agichtein}]{10.1145/3331184.3331375}
Ali Ahmadvand, Jason~Ingyu Choi, and Eugene Agichtein. 2019.
\newblock \href {https://doi.org/10.1145/3331184.3331375} {Contextual dialogue
  act classification for open-domain conversational agents}.
\newblock In \emph{Proceedings of the 42nd International ACM SIGIR Conference
  on Research and Development in Information Retrieval}, SIGIR'19, page
  1273^^e2^^80^^931276, New York, NY, USA. Association for Computing Machinery.

\bibitem[{Devlin et~al.(2019)Devlin, Chang, Lee, and
  Toutanova}]{devlin-etal-2019-bert}
Jacob Devlin, Ming-Wei Chang, Kenton Lee, and Kristina Toutanova. 2019.
\newblock \href {https://doi.org/10.18653/v1/N19-1423} {{BERT}: Pre-training of
  deep bidirectional transformers for language understanding}.
\newblock In \emph{Proceedings of the 2019 Conference of the North {A}merican
  Chapter of the Association for Computational Linguistics: Human Language
  Technologies, Volume 1 (Long and Short Papers)}, pages 4171--4186,
  Minneapolis, Minnesota. Association for Computational Linguistics.

\bibitem[{Fan et~al.(2018)Fan, Lewis, and Dauphin}]{fan-etal-2018-hierarchical}
Angela Fan, Mike Lewis, and Yann Dauphin. 2018.
\newblock \href {https://doi.org/10.18653/v1/P18-1082} {Hierarchical neural
  story generation}.
\newblock In \emph{Proceedings of the 56th Annual Meeting of the Association
  for Computational Linguistics (Volume 1: Long Papers)}, pages 889--898,
  Melbourne, Australia. Association for Computational Linguistics.

\bibitem[{Ghazarian et~al.(2019)Ghazarian, Wei, Galstyan, and
  Peng}]{ghazarian-etal-2019-better}
Sarik Ghazarian, Johnny Wei, Aram Galstyan, and Nanyun Peng. 2019.
\newblock \href {https://doi.org/10.18653/v1/W19-2310} {Better automatic
  evaluation of open-domain dialogue systems with contextualized embeddings}.
\newblock In \emph{Proceedings of the Workshop on Methods for Optimizing and
  Evaluating Neural Language Generation}, pages 82--89, Minneapolis, Minnesota.
  Association for Computational Linguistics.

\bibitem[{Jiang and de~Rijke(2018)}]{jiang-de-rijke-2018-sequence}
Shaojie Jiang and Maarten de~Rijke. 2018.
\newblock \href {https://doi.org/10.18653/v1/W18-5712} {Why are
  sequence-to-sequence models so dull? understanding the low-diversity problem
  of chatbots}.
\newblock In \emph{Proceedings of the 2018 {EMNLP} Workshop {SCAI}: The 2nd
  International Workshop on Search-Oriented Conversational {AI}}, pages 81--86,
  Brussels, Belgium. Association for Computational Linguistics.

\bibitem[{Kawano et~al.(2019)Kawano, Yoshino, and
  Nakamura}]{kawano-etal-2019-neural}
Seiya Kawano, Koichiro Yoshino, and Satoshi Nakamura. 2019.
\newblock \href {https://doi.org/10.18653/v1/W19-8627} {Neural conversation
  model controllable by given dialogue act based on adversarial learning and
  label-aware objective}.
\newblock In \emph{Proceedings of the 12th International Conference on Natural
  Language Generation}, pages 198--207, Tokyo, Japan. Association for
  Computational Linguistics.

\bibitem[{Liu et~al.(2016)Liu, Lowe, Serban, Noseworthy, Charlin, and
  Pineau}]{liu-etal-2016-evaluate}
Chia-Wei Liu, Ryan Lowe, Iulian Serban, Mike Noseworthy, Laurent Charlin, and
  Joelle Pineau. 2016.
\newblock \href {https://doi.org/10.18653/v1/D16-1230} {How {NOT} to evaluate
  your dialogue system: An empirical study of unsupervised evaluation metrics
  for dialogue response generation}.
\newblock In \emph{Proceedings of the 2016 Conference on Empirical Methods in
  Natural Language Processing}, pages 2122--2132, Austin, Texas. Association
  for Computational Linguistics.

\bibitem[{Papineni et~al.(2002)Papineni, Roukos, Ward, and
  Zhu}]{papineni-etal-2002-bleu}
Kishore Papineni, Salim Roukos, Todd Ward, and Wei-Jing Zhu. 2002.
\newblock \href {https://doi.org/10.3115/1073083.1073135} {{B}leu: a method for
  automatic evaluation of machine translation}.
\newblock In \emph{Proceedings of the 40th Annual Meeting of the Association
  for Computational Linguistics}, pages 311--318, Philadelphia, Pennsylvania,
  USA. Association for Computational Linguistics.

\bibitem[{Raffel et~al.(2020)Raffel, Shazeer, Roberts, Lee, Narang, Matena,
  Zhou, Li, and Liu}]{2020t5}
Colin Raffel, Noam Shazeer, Adam Roberts, Katherine Lee, Sharan Narang, Michael
  Matena, Yanqi Zhou, Wei Li, and Peter~J. Liu. 2020.
\newblock \href {http://jmlr.org/papers/v21/20-074.html} {Exploring the limits
  of transfer learning with a unified text-to-text transformer}.
\newblock \emph{Journal of Machine Learning Research}, 21(140):1--67.

\bibitem[{Raheja and Tetreault(2019)}]{raheja-tetreault-2019-dialogue}
Vipul Raheja and Joel Tetreault. 2019.
\newblock \href {https://doi.org/10.18653/v1/N19-1373} {{D}ialogue {A}ct
  {C}lassification with {C}ontext-{A}ware {S}elf-{A}ttention}.
\newblock In \emph{Proceedings of the 2019 Conference of the North {A}merican
  Chapter of the Association for Computational Linguistics: Human Language
  Technologies, Volume 1 (Long and Short Papers)}, pages 3727--3733,
  Minneapolis, Minnesota. Association for Computational Linguistics.

\bibitem[{Roller et~al.(2021)Roller, Dinan, Goyal, Ju, Williamson, Liu, Xu,
  Ott, Smith, Boureau, and Weston}]{roller-etal-2021-recipes}
Stephen Roller, Emily Dinan, Naman Goyal, Da~Ju, Mary Williamson, Yinhan Liu,
  Jing Xu, Myle Ott, Eric~Michael Smith, Y-Lan Boureau, and Jason Weston. 2021.
\newblock \href {https://doi.org/10.18653/v1/2021.eacl-main.24} {Recipes for
  building an open-domain chatbot}.
\newblock In \emph{Proceedings of the 16th Conference of the European Chapter
  of the Association for Computational Linguistics: Main Volume}, pages
  300--325, Online. Association for Computational Linguistics.

\bibitem[{Shriberg et~al.(2004)Shriberg, Dhillon, Bhagat, Ang, and
  Carvey}]{shriberg-etal-2004-icsi}
Elizabeth Shriberg, Raj Dhillon, Sonali Bhagat, Jeremy Ang, and Hannah Carvey.
  2004.
\newblock \href {https://aclanthology.org/W04-2319} {The {ICSI} meeting
  recorder dialog act ({MRDA}) corpus}.
\newblock In \emph{Proceedings of the 5th {SIG}dial Workshop on Discourse and
  Dialogue at {HLT}-{NAACL} 2004}, pages 97--100, Cambridge, Massachusetts,
  USA. Association for Computational Linguistics.

\bibitem[{Stolcke et~al.(2000)Stolcke, Ries, Coccaro, Shriberg, Bates,
  Jurafsky, Taylor, Martin, Van Ess-Dykema, and
  Meteer}]{stolcke-etal-2000-dialogue}
Andreas Stolcke, Klaus Ries, Noah Coccaro, Elizabeth Shriberg, Rebecca Bates,
  Daniel Jurafsky, Paul Taylor, Rachel Martin, Carol Van Ess-Dykema, and Marie
  Meteer. 2000.
\newblock \href {https://aclanthology.org/J00-3003} {Dialogue act modeling for
  automatic tagging and recognition of conversational speech}.
\newblock \emph{Computational Linguistics}, 26(3):339--374.

\bibitem[{Zhao et~al.(2017)Zhao, Zhao, and Eskenazi}]{zhao-etal-2017-learning}
Tiancheng Zhao, Ran Zhao, and Maxine Eskenazi. 2017.
\newblock \href {https://doi.org/10.18653/v1/P17-1061} {Learning
  discourse-level diversity for neural dialog models using conditional
  variational autoencoders}.
\newblock In \emph{Proceedings of the 55th Annual Meeting of the Association
  for Computational Linguistics (Volume 1: Long Papers)}, pages 654--664,
  Vancouver, Canada. Association for Computational Linguistics.

\bibitem[{Zhao et~al.(2020)Zhao, Lala, and Kawahara}]{zhao-etal-2020-designing}
Tianyu Zhao, Divesh Lala, and Tatsuya Kawahara. 2020.
\newblock \href {https://doi.org/10.18653/v1/2020.acl-main.4} {Designing
  precise and robust dialogue response evaluators}.
\newblock In \emph{Proceedings of the 58th Annual Meeting of the Association
  for Computational Linguistics}, pages 26--33, Online. Association for
  Computational Linguistics.

\end{thebibliography}
\bibliographystyle{acl_natbib}

\end{document}